\pdfoutput=1

\documentclass[11pt]{article}

\usepackage[final]{acl}

\usepackage{times}
\usepackage{latexsym}

\usepackage[T1]{fontenc}

\usepackage[utf8]{inputenc}

\usepackage{microtype}

\usepackage{inconsolata}

\usepackage{graphicx}
\usepackage{booktabs}
\usepackage{enumitem}
\usepackage{color}

\usepackage{arydshln}
\usepackage{svg}
\usepackage{multirow}
\usepackage{subfigure}

\title{No Train but Gain:\\Language Arithmetic for training-free Language Adapters enhancement}

\author{
 \textbf{Mateusz Klimaszewski\textsuperscript{1,2}},
 \textbf{Piotr Andruszkiewicz\textsuperscript{1}},
 \textbf{Alexandra Birch\textsuperscript{2}}
 \\
 \textsuperscript{1}Warsaw University of Technology,
 \textsuperscript{2}University of Edinburgh
\\
 \small{
   \textbf{Correspondence:} \href{mailto:email@domain}{mklimasz@ed.ac.uk}
 }
}

\begin{document}
\maketitle
\begin{abstract}
Modular deep learning is the state-of-the-art solution for lifting the curse of multilinguality, preventing the impact of negative interference and enabling cross-lingual performance in Multilingual Pre-trained Language Models. However, a trade-off of this approach is the reduction in positive transfer learning from closely related languages. In response, we introduce a novel method called language arithmetic, which enables training-free post-processing to address this limitation.
Extending the task arithmetic framework, we apply learning via addition to the language adapters,  transitioning the framework from a multi-task to a multilingual setup. The effectiveness of the proposed solution is demonstrated on three downstream tasks in a MAD-X-based set of cross-lingual schemes, acting as a post-processing procedure. Language arithmetic consistently improves the baselines with significant gains, especially in the most challenging case of zero-shot application. Our code and models are available at \url{https://github.com/mklimasz/language-arithmetic}.
\end{abstract}

\section{Introduction}
The recent progress of large language models has raised the question of how well they  perform not just in English but across multiple languages which has spurred interest in Multilingual Pre-trained Language Models (MLLMs) \cite{conneau-etal-2020-unsupervised,workshop2023bloom,alves2024tower}.
These models serve as general-purpose solutions that can be adapted and applied to various Natural Language Processing tasks. Notably, MLLMs demonstrate zero-shot cross-lingual capabilities, allowing them to generalise effectively to downstream tasks even when pre-trained in a language different from the target language.

The positive transfer of abilities from both related languages and high-quality training data from unrelated languages has meant that MLLMs have reported state-of-the-art performance in low-resourced languages \cite{muennighoff-etal-2023-crosslingual}.
However, this benefit does not always extend to high-resourced languages \cite{kocmi2022reality}. In such cases, the quality of MLLMs tends to decrease compared to their monolingual counterparts \cite[among others]{Nozza2020WhatT,martin-etal-2020-camembert} due to negative interference phenomena~\cite{wang-etal-2020-negative}. Additionally, the curse of multilinguality \cite{conneau-etal-2020-unsupervised} reveals the existence of a trade-off between language coverage and model capacity. Consequently, MLLMs must carefully limit the number of languages included during the pre-training phase.

Modular deep learning (MDL) \cite{pfeiffer2023modular} methods help to avoid negative interference and limited model capacity
, enabling the extension of MLLMs to support any number of languages
\cite{bapna-firat-2019-simple,ustun-etal-2020-udapter,philip-etal-2020-monolingual,pfeiffer-etal-2020-mad,pfeiffer-etal-2022-lifting}.
MDL methods adapt the model to arbitrary tasks and languages by isolating components from each other (and the backbone MLLM) via parameter-efficient extensions.
Examples of parameter-efficient modules are
adapters \cite{NIPS2017_e7b24b11,pmlr-v97-houlsby19a}, which are low-budget (in terms of parameters) bottleneck layers that increase an MLLM size by just a fraction. Language adapters \cite{pfeiffer-etal-2020-mad} allow the modularisation of language-specific knowledge by training on a raw, unlabelled corpus for specific languages. 

The limitation of the MDL and language adapters is their isolation. While they lift the curse of multinguality and prevent negative interference, at the same time, language adapters limit the possible impact of positive transfer. Previous attempts to address these challenges — such as training bilingual \cite{parovic-etal-2022-bad} or language-family \cite{chronopoulou-etal-2023-language} adapters — do not scale effectively. In our work, we tackle this limitation as a post-processing step. Leveraging recent insights from task arithmetic \cite{DBLP:conf/iclr/IlharcoRWSHF23}, specifically \textit{learning via addition}, we augment language adapters with missing related language knowledge -- a concept we term \textbf{language arithmetic}. Remarkably, this training-free approach can enhance not only existing language adapters but also offer zero-shot performance.

To summarise, our contributions are as follows:

\begin{itemize}
    \item A novel training-free post-processing method named language arithmetic that enhances language adapters. 
    \item We conduct a cross-lingual evaluation on three downstream tasks (NER, NLI and QA) and two Multilingual Pre-trained Language Models (XLM-R, mBERT) with test cases that include zero-shot and low-resource setups in a diverse group of 13 languages.
    \item We provide an analysis of language arithmetic internal components (including a comparison with task arithmetic) and show improvement up to 3 F1 points without any additional training involved.
\end{itemize}

\section{Background}
Our research builds upon the task arithmetic contributions of \citet{DBLP:conf/iclr/IlharcoRWSHF23} and \citet{zhang2023composing}. The following Section provides the background and serves as a gentle introduction to the concept of task vectors and task arithmetic.

\subsection{Task vectors \& Task arithmetic}
Let us assume that we have access to a pre-trained model with its weights denoted $\theta_{pre} \in R^d$  and a fine-tuned version of the same model on a task $t$ represented by $\theta^{t}_{ft} \in R^d$. The task vector $\tau_{t} \in R^d$ is an element-wise difference between models' weights.
\begin{equation}
    \tau_{t} = \theta^{t}_{ft} - \theta_{pre}
\end{equation}
The task vectors can be part of multiple arithmetic operations, e.g. \textit{learning via addition}. This operation is an addition operation between two task vectors and the base model, i.e. we add two differences between the fine-tuned models and the pre-trained version with weights controlling the impact.
\begin{equation}
    \theta_{multi-task} = \theta_{pre} + \lambda_1 \tau_{t_1} + \lambda_2 \tau_{t_2}
\end{equation}
The lambdas can be further normalised to sum to one, i.e. $\lambda_2 = 1 - \lambda_1$ and simplifying notation with just~$\lambda$.
\begin{equation}
    \theta_{multi-task} = \theta_{pre} + \lambda \tau_{t_1} + (1 - \lambda) \tau_{t_2}
\end{equation}
While we define learning via addition for two tasks, the same procedure can be applied to multiple tasks.

Task arithmetic allows us to forge a multi-task model from a separate, task-specific set of fine-tuned models, preserving high accuracy (although a shared pre-trained starting point is required, e.g. the same Language Model). Moreover, vectors from different tasks are typically close to orthogonal, and \citet{DBLP:conf/iclr/IlharcoRWSHF23} speculate that this enables the combination of task vectors via addition with minimal interference.

In our work, we focus on parameter-efficient fine-tuning (PEFT) and use language adapters. Therefore, we reduce the task vector and underlying model weights represented by $\theta$ to newly introduced parameters (i.e. we exclude the backbone MLLM, which is frozen across all the models, following the work of \citet{zhang2023composing})

\section{Method}

We propose language arithmetic that transitions the task arithmetic concept from a multi-task to a multilingual setup. In this Section, we describe the language arithmetic alongside its application as a training-free, post-processing step to a MAD-X cross-lingual framework \cite{pfeiffer-etal-2020-mad}.

\subsection{Language arithmetic}
We formulate a language arithmetic (LA) concept by substituting the task in task vectors and arithmetic with a language. This approach means that instead of merging downstream tasks, we target a problem of cross-lingual performance. We propose to apply \textit{learning via addition} to languages, and in Appendix \ref{sec:lang_vs_task}, we demonstrate the discrepancies when comparing language and task vectors. Our study focuses specifically on the language adapters \cite{philip-etal-2020-monolingual,pfeiffer-etal-2020-mad}. Due to overlapping abbreviations, we use the LA exclusively as the former, i.e. language arithmetic. In the learning via addition, we limit the parameters to language adapters and simplify the notation that $\theta$ represents the adapters' weights and $\tau$ is referred to as a language vector. As we operate in a language space instead of a task, the $t$ is replaced with a language, i.e. its language code in the notation. The example equation describes a language arithmetic operation between an English and a Spanish adapter.
\begin{equation}
    \theta_{LA} = \theta_{pre} + \lambda \tau_{en} + (1 - \lambda) \tau_{es}
\end{equation}
Throughout the paper, the equation above is abbreviated as a function: $LA(en, es)$ with lambda as a default parameter. Additionally, for clarity reasons, we denote target language in a subscript to distinguish different use-cases. For example, $LA_{fr}(en, es)$ means that language arithmetic between English and Spanish is applied to a different language - in this case French (zero-shot application), or $LA_{es}(en, es)$ meaning that the target is Spanish (non-zero-shot use-case, due to Spanish being a part of the LA equation).

\begin{figure*}[t]
    \centering
    \includesvg[width=0.7\linewidth]{figures/LA-train.drawio.svg}
    \caption{
    Language arithmetic as an extension of the MAD-X framework. Given language and task adapters (left), language arithmetic (right) enables post-processing, training-free improvement in two use-cases: (i) zero-shot where a language adapter for a target language was not trained (presented in the figure as Spanish, which was not part of existing language adapters pool, $LA_{es}(en, fr)$) or (ii) to improve existing language adapters via arithmetic with either related language or a language on which task adapter was trained (e.g. $LA_{fr}(en, fr)$).   
    }
    
    \label{fig:overview}
\end{figure*}

Language arithmetic is a training-free method, taking advantage of already pre-trained modules. The sole requirement is a validation dataset on which the $\lambda$ parameter can be established. While in our work, we use a pretty fine-grained step ($0.05$) to determine the $\lambda$ value (i.e. we run evaluations for $\lambda \in [0, 1]$ with a provided step), our analysis showcased that it is possible to increase the value and limit the computation burden even more (details in Section \ref{sec:analysis_lambda}).

\subsection{Application}

We evaluate our post-processing method as an extension of the MAD-X framework \cite{pfeiffer-etal-2020-mad} to challenge our solution in a cross-lingual manner. The overview of the schema is presented in Figure \ref{fig:overview}.

The MAD-X consists of the following steps:
\begin{enumerate}
    \item Training language adapter(s)
    \item Training task adapter
    \item Cross-lingual inference
\end{enumerate}

\noindent In our work, we introduce an additional step:
\begin{enumerate}\addtocounter{enumi}{3}
    \item Post-processing via language arithmetic
\end{enumerate}

In the following Sections \ref{sec:training}-\ref{sec:post_processing}, we present the framework and our proposed post-processing via language arithmetic extension, exploring two use cases: (i) a zero-shot case, where a target language adapter does not exist and (ii) an enhancement case, where we prove existing language adapters (in high- and low-resourced languages).

\subsubsection{Training language adapter(s)}
\label{sec:training}
In the first step, the MAD-X framework trains language adapters. These adapters are trained on raw corpora using masked language modelling loss in a self-supervised manner. The MLLM is frozen during this step, and we only optimise the newly introduced adapter. The training must be done for languages corresponding to the downstream tasks (e.g. if we have an English NER dataset, we need an English language adapter, apart from other desired target languages). Additionally, the adapters form a pool that is leveraged during cross-lingual inference.

\subsubsection{Training task adapters}
The following step freezes a backbone MLLM and a language adapter and trains a task adapter on a downstream task dataset. Given a set of tasks or if a new task appears, we can repeat this step as long as the required language adapter exists in the available pool, i.e. a language adapter that matches the task's language.

\subsubsection{Cross-lingual inference}
Having trained a task adapter, we can leverage a pool of pre-trained language adapters and obtain a cross-lingual performance by connecting any existing language adapter with a newly trained task adapter (i.e. routing first via language adapter and then task adapter). The growing pool of pre-trained adapters can be accessed at public repositories like AdapterHub \cite{pfeiffer-etal-2020-adapterhub} and reused for further use cases.

\subsubsection{Post-processing via language arithmetic}
\label{sec:post_processing}
Our method builds upon the MAD-X framework in two enhancement scenarios.

First, we assume a situation where the pool of language adapters does not contain a desired target language, i.e. a zero-shot scenario. In contrast to the previous works that try to improve existing adapters, this use-case is an alternative to routing via either a related language or a task language (here, by task language, we understand the language on which the task adapter was trained, in contrast to a target language - on which we want to evaluate). Instead of choosing the better-performing proxy, language arithmetic proposes to combine these two (with better results, as shown in Section \ref{sec:zero_shot}).

In the second language adapter enhancement scenario, we apply language arithmetic as a more common goal, trying to improve existing language adapters; however, our method does that without any training. Here, we combine the existing target language adapter with either a related language (we define related languages in Section \ref{para:related}) or, once again, a task language.

\section{Experiments}

\subsection{Experimental setup}

\subsubsection{Datasets}
Downstream evaluation is performed on three tasks: Named Entity Recognition (NER), Natural Language Inference (NLI) and Question Answering (QA), covering jointly 13 languages\footnote{ar, bg, de, el, es, fr, hi, ru, sw, tr, ur, vi, zh; XQuAD does not cover 4 languages (bg, fr, sw, ur)}, while the training - to perform cross-lingual evaluation - is performed on the English data. For the NER task, we use the WikiANN \cite{rahimi-etal-2019-massively} dataset and for NLI - XNLI \cite{conneau2018xnli}. The QA evaluation is done on XQuAD \cite{artetxe-etal-2020-cross} (we split data 50/50 into valid/test datasets), and the training uses SQuAD 1.1 \cite{rajpurkar-etal-2016-squad}. Additionally, to evaluate a low-resource scenario for a language not covered during MLLM pre-training, we leverage the Assamese subset from IndicXNLI \cite{aggarwal-etal-2022-indicxnli}.

\paragraph{Related languages}\label{para:related} To automatically establish a related language needed for language arithmetic, we query URIEL and lang2vec library \cite{littell-etal-2017-uriel,malaviya-etal-2017-learning}. During the related language query, we limited the options to 13 downstream task languages for which we had already pre-computed language adapters. This limitation means that for some languages, we would be able to find a stronger performing pairing and that the ceiling for our method is higher than the presented (we denote that the performance of hypothetical paring would also depend on data availability, i.e. a paired language must be not only related but also have representative corpora; based on this we show analysis and improved performance in Section \ref{sec:romance}). However, considering the limitations of the lang2vec, we decided to keep this simplification. At last, a language can have a set of equally good related languages. Therefore, in practical terms, it is not feasible for our study to train all possible options for each language - our simplification stands as a reasonable, real-world proxy. We provide the list of related languages in the Appendix \ref{sec:appendix_related}.

\subsubsection{Implementation \& training}
In our work, we focus on two of the most popular multilingual PLMs\footnote{According to downloads from Huggingface \url{hf.co/models?language=multilingual&sort=downloads}}: mBERT\footnote{\texttt{bert-base-multilingual-cased}} \cite{devlin-etal-2019-bert} and XLM-R\footnote{\texttt{xlm-roberta-base}} \cite{conneau-etal-2020-unsupervised}. We implement our method using the AdapterHub library \cite{pfeiffer-etal-2020-adapterhub}. For language adapters, we train on the Wikipedia corpora\footnote{20331101.xx checkpoint \url{hf.co/datasets/wikimedia/wikipedia}} for 250k steps with a learning rate of 1e-4, an effective batch size of 64 using a single GPU
and the same initialisation. For task-specific training, we train for 100 epochs with the same learning rate and a batch size set to 16. We choose the final checkpoint based on validation dataset performance  (for language adapters, we evaluate on a held-out subset of Wikipedia). In our main experiments, we report the scores as an average over three independent runs with different seeds (for both language and task adapters).
Additionally, to improve efficiency and reduce GPU memory utilisation, we adopt a half-precision (FP16) setting.

\subsection{Zero-shot evaluation}
\label{sec:zero_shot}

The zero-shot evaluation assumes a scenario where the language adapter pool does not contain a desired target language (e.g. lack of Spanish in Figure \ref{fig:overview}). The baselines are based on routing, i.e. we proxy either by an English adapter (proxy via a task language, as the task adapter was trained using English data) or a related language (e.g. French for Spanish). Language arithmetic serves a solution that, instead of choosing a better proxy, combines the adapter's tuple: $LA_{t}(en, rel)$, where $rel$ symbolises a related language and $t$ stands for a target language (e.g. $LA_{es}(en, fr)$).

\begin{figure}[t]
    \centering
    \includesvg[width=0.88\linewidth]{figures/NER_XLMR_zeroshot.svg}

    \vspace{-0.25cm}

    \includesvg[width=0.88\linewidth]{figures/NLI_XLMR_zeroshot.svg}

    \vspace{-0.25cm}

    \includesvg[width=0.88\linewidth]{figures/QA_XLMR_zeroshot.svg}
    
    \caption{Zero-shot XLM-R language arithmetic evaluation, where one side of the arithmetic is an English adapter, and the other is related to the target language adapter (e.g. French for Spanish - $LA_{es}(en, fr)$). The values above bars present a relative difference to a better proxy. See Figure \ref{fig:mBERT_zeroshot} for the mBERT model.}
    \label{fig:xlmr_zeroshot}
\end{figure}

Figures \ref{fig:xlmr_zeroshot} and \ref{fig:mBERT_zeroshot} present the results of the zero-shot experiment. Language arithmetic consistently outperforms the proxy baselines for all the setups, reaching over $3.1$ F1 points improvement in the NER task and $1.1$ F1 for QA (XLM-R). These results indicate that language arithmetic is a feasible, low-cost method that one can apply in the lack of an existing target language adapter.

Additionally, we investigate how the $\lambda$ parameter impacts the downstream evaluation. The goal was to understand how much weight is given to English vs related language. We looked at the validation performance over different $\lambda$ thresholds. While most cases set the value to over $0.5$ (i.e. preferring the English side, given $LA_{t}(en, rel)$), the preferred values did not showcase any consistency and pattern. We analyse this further in Section \ref{sec:analysis_lambda}.

\subsection{Improving existing language adapters}
\label{sec:improving}

This evaluation assumes that a target language adapter exists in the adapter pool. We test two cases, i.e. $LA_{t}(en, t)$ and $LA_{t}(rel, t)$, where $rel$ is once again a related language and $t$ is the target language. Additionally, we provide a combination of these two approaches (referred to as $LA_{t}(en/rel, t)$), where for each language, we choose a better pairing (so either $en$ or $rel$). This solution resembles a practical compromise between cost and performance and serves as a proxy for the ceiling of our method (discussed in Section \ref{para:related}).

The results are presented in Figures \ref{fig:xlmr_improving} and \ref{fig:mbert_improving}. Compared to the baseline direct application of a target language adapter (i.e., the MAD-X method), the gains are not as significant as in the case of the zero-shot scenario. Moreover, in contrast to the previous Section's study, MLLMs showcase a different behaviour, as language arithmetic provides less benefit for XLM-R than for the mBERT model (e.g. $+0.38$ XLM-R vs $+2.41$ mBERT in F1, QA).

The drop in performance of language arithmetic compared to the zero-shot use case is not surprising. Given that a target language adapter is trained on a significant corpus, it gives less room for improvement (this is not the case in the low-resource regimes, as shown in the following Section). This is also a potential explanation of a different behaviour between the evaluated MLLMs, considering the overall more robust performance of XLM-R over mBERT. However, considering the cost-to-performance ratio and the minimal fatigue that our post-processing method enforces on existing MAD-X pipelines, we can see a constant gain on average across all the experiments and training runs.
\begin{figure}[t]
    \centering
    \includesvg[width=0.88\linewidth]{figures/NER_XLMR.svg}

    \vspace{-0.25cm}

    \includesvg[width=0.88\linewidth]
    {figures/NLI_XLMR.svg}

    \vspace{-0.25cm}

    \includesvg[width=0.88\linewidth]{figures/QA_XLMR.svg}
    
    \caption{Variants of language arithmetic compared to the MAD-X method in the use-case to improve an existing target language adapter. The values above bars present a difference between a better LA setup and the MAD-X framework for the XLM-R model (see Figure \ref{fig:mbert_improving} for mBERT).}
    \label{fig:xlmr_improving}
\end{figure}

\subsection{Low resource evaluation}

\begin{figure*}[t]
  \centering
  \hspace{0.4cm}
 \subfigure[NER]{\includesvg[width=0.43\linewidth]{figures/low_resource_ner}}\hfill
  \subfigure[NLI]{\includesvg[width=0.43\linewidth]{figures/low_resource_nli}}
  \hspace{0.4cm}
\caption{NER and NLI evaluation of a set of adapters trained on a Wikipedia subset showcases that language arithmetic $LA_t(t,en)$ (green, dotted line) provides significant gains when compared against direct usage of the adapter (violet, solid line), especially in a very low-resource regime. The x-axis represents the token budget of each trained language adapter.}
\label{fig:lowres_ner_nli}
\end{figure*}

Training a language adapter might be troublesome for high-resourced languages due to massive corpora requiring significant computational resources.\footnote{Although in our experimental setup, we train each adapter for the same number of steps and choose the best checkpoint based on the validation performance, for low-resourced languages, one could apply an early stopping mechanism in a production-level pipeline.} On the other hand, in most languages, we lack data to train a strong language adapter, i.e. language-specific corpora might be either too small or unavailable  \cite{haddow-etal-2022-survey}. We investigate whether LA can help in such cases. We test our solution in three cases and define three (actual and simulated) evaluation scenarios:
\begin{itemize}
    \item Assamese (as) - low resource language, additionally not used in the pre-training of a base MLLM,
    \item Swahili (sw) - low resource language, used in the pre-training,
    \item French (fr) - high resource, used in the pre-training. We simulate cases from low to high resources.
\end{itemize}
We train a series of language adapters with different token budgets for each language, from 10k to 10M (or 1B for French; we limit this particular study to the XLM-R model). Afterwards, we compare the usage of such adapters directly against language arithmetic with three adapters (we use $LA_t(t,en)$, where $t \in \{as,sw,fr\}$).

Figure \ref{fig:lowres_ner_nli} presents the results of the evaluation performed on the downstream tasks. The most gain is visible in the most challenging scenario, during the evaluation on the Assamese dataset. In this case, the backbone MLLMs did not encounter the language during the pre-training phase. Although the difference becomes less pronounced in the NER task as we approach the limits of available data, there remains a significant margin for NLI - the difference can be explained by the overlap in the corpora (Wikipedia) between NER and language adapter training tasks, following the findings of \citet{gururangan-etal-2020-dont}. For Swahili, where the language is part of the pre-training, the flattening effect begins earlier and affects both tasks. Nevertheless, leveraging language arithmetic still yields improvements.

The simulated case of French showcases that even with a relatively weak language adapter (trained on 10k tokens), the language arithmetic can restore existing knowledge and results in high performance for the language. Moreover, comparing the adapters trained with a different token budget, the results remain similar, without significant fluctuations.  We believe that this phenomenon happens because the MLLM has seen a much higher amount of French in the pre-training procedure than Swahili (over $35$ times more tokens in XLM-R pre-training; moreover, French is in the top $15$ represented languages). Therefore, even undertrained French adapters have a relatively easy task once they are merged with a robust English adapter. In practical terms, this finding allows us to prototype new languages quicker by estimating the possible end product quality or might serve as an intermediate solution (until the full-corpora adapter is trained).

\section{Analysis}

\subsection{Lambda impact}
\label{sec:analysis_lambda}
Our study estimates the $\lambda$ parameter with a small step ($0.05$). This analysis investigates how sensitive this parameter is in the language arithmetic. Depending on multiple variables that include model and evaluation dataset sizes or a number of languages, running $20$ evaluations might be costly (especially when using neural-based metrics, e.g. COMET \cite{rei-etal-2020-comet}). Therefore, we analysed the potential impact of choosing a suboptimal lambda with a decrease in evaluation count. The breakdown includes a subset of languages on both tasks (using the XLM-R as a base model). We chose the zero-shot scenario where we performed LA between English and related language adapters.

In Figures \ref{fig:zeroshot_interpolation} and \ref{fig:zeroshot_interpolation_nli_qa}, we plot the validation scores with the corresponding baselines, that is, the scores of using directly the adapters. The dotted lines are based on $\lambda=0$ or $\lambda=1$ for clarity, meaning we exclusively use the arithmetic equation's left or right side (i.e., a specific language). In most cases, a subset of valid $\lambda$ values would improve over the baselines. Moreover, the analysis reveals that a coarser evaluation (e.g., with a step of $0.1$ or $0.2$) would be sufficient, reducing the required number of performed tests up to four times while maintaining most of the improvement. At last, setting the default $\lambda=0.5$ would be near optimum for the analysed subset.

\begin{figure}[t]
  \centering
  \includesvg[width=0.9\linewidth]{figures/zeroshot_ner_xlmr}
\caption{Interpolation of $\lambda$ values for the zero-shot XLM-R scenario (NER, for NLI and QA see Appendix \ref{sec:appendix_lambda_nli_qa}) on the validation dataset. The horizontal dashed lines represent the baseline scores for both languages used in language arithmetic.}
\label{fig:zeroshot_interpolation}
\end{figure}

\subsection{Language relatedness}
\label{sec:romance}

\begin{table*}[th!]
    \centering
    \begin{tabular}{llcccccc}
        \toprule
        \multicolumn{2}{c}{$LA_{l_1}(l_1 \downarrow,l_2 \rightarrow)$} & ca & es & fr & it & pt & ro \\
        \midrule
        \multirow{6}{*}{\rotatebox[origin=c]{90}{Eval language}} &&  \multicolumn{6}{c}{NER} \\
        & es & 73.82 & - & 75.06 & 74.61 & 74.90 & 73.68 \\
        & fr & 75.74 & 75.76 & - & 75.82 & 75.59 & 75,79 \\
        \cdashline{2-8}
        && \multicolumn{6}{c}{NLI} \\
        & es & 78.23 & - & 78.04 & 77.96 & 77.94 & 78,02 \\
        & fr & 77.95 & 77,65 & - & 77.70 & 77,47 & 77,60 \\
        \bottomrule
    \end{tabular}
    \caption{\label{tab:related_impact}Impact of language relatedness on the language arithmetic. We compare different Romance languages as a right side of $LA$ equation, i.e. $l2$ (both tasks use XLM-R model). We report an average over three runs.}
\end{table*}

Relatedness of languages is a difficult-to-define concept. At times, in our proposed framework, we might face a choice of multiple, seemingly equally related languages to use for the arithmetic operation. In this analysis, we decided to look at this aspect via a glance at Romance languages. We trained an additional subset of language adapters and formed a pool of 6 languages: Catalan, French, Italian, Portuguese, Romanian and Spanish. Afterwards, we evaluated languages shared in our NER and NLI tasks (Spanish and French) by arithmetic with the entire Romance languages pool. 

The results are presented in Table \ref{tab:related_impact} and show that given a different related language (in this case, defined as coming from the same language family), there are minor scores fluctuation. The relative difference between the best and the worst language reaches around $1$ F1 score in the NER task and around $0.3$ in terms of accuracy points for the NLI task. This experiment indicates that a more sophisticated or hand-crafted language choice would improve the downstream results presented in Figure 5. However, it also shows that there is no free lunch, and results depend on a downstream task. For example, for Spanish evaluation, a French adapter is trained on the largest out of the listed languages raw corpora; therefore, for the NER task, it can leverage a bigger pool of seen during training Named Entities (at times language-independent or similar across languages) and perform the best given more data, even when there are closer related languages (according to methodology from Section \ref{para:related} and Indo-European languages family tree).

\section{Related Work}
Knowledge composition from multiple, independently trained adapters has been widely discussed in the literature. However, unlike our work, the solutions require substantial changes to the vanilla adapter setup. The previous work either requires additional parameters to learn a parameterised composition function/a gating module to combine/steer the flow through the suitable adapter(s), or needs a specific training procedure that increases the complexity of the overall solution or, in most cases, both \cite{pham-etal-2020-study,pfeiffer-etal-2021-adapterfusion,lee-etal-2022-fad,parovic-etal-2022-bad,chronopoulou-etal-2023-language,klimaszewski23,wang-etal-2023-adapterdistillation}. Moreover, to prevent specifically negative interference, hyper-adapters \cite{baziotis-etal-2022-multilingual} were proposed using hyper-networks \cite{ha2017hypernetworks}, and \citet{ansell-etal-2022-composable} applied sparse fine-tuning to compose task and language masks. Unlike the prior studies mentioned earlier, our attempt is training-free and does not modify the base architecture. The most conceptually similar work is proposed by \citet{chronopoulou-etal-2023-adaptersoup}; however, they operate on the notion of sample similarity to a subset of domains in a domain adaptation regime. Additionally, concurrent to our work, \citet{parovic-etal-2024-investigating} show intial potential of task arithemtic in cross-lingual transfer based on a full fine-tuning setup. However, in our work we focus on PEFT methods with additional, in-depth analysis. At last, we denote the rise of task arithmetic use cases, e.g. vision tasks or cross-task generalisation \cite{stoica2023zipit,huang2024lorahub}.

\section{Conclusion}

We have proposed language arithmetic, which enhances language adapters based on task arithmetic learning via addition. It is a training-free method and functions as a post-processing technique for MAD-X. Our experiments have shown that LA is particularly beneficial in a zero-shot scenario, where we do not have access to a target language adapter. At last, we highlight the differences between language and task arithmetic.

In our future work, we plan to extend language arithmetic by incorporating more components into the sum. Additionally, we aim to adapt other elements of the task arithmetic framework, i.e. task analogies and forgetting via negation, to a multilingual setup with an analysis of the differences between multi-task and multilingual arithmetic context. Furthermore, we will evaluate LA's performance on various non-classification tasks.

\section{Limitations}
Our work was tested on English-centric task training and could be extended to different languages with more PEFT methods. Moreover, applying multi-source training based on the work of \citet{ansell-etal-2023-unifying} could provide better robustness of the task adapters and a more thorough analysis.

\bibliography{latex/acl_latex}

\begin{thebibliography}{46}
\providecommand{\natexlab}[1]{#1}

\bibitem[{Aggarwal et~al.(2022)Aggarwal, Gupta, and Kunchukuttan}]{aggarwal-etal-2022-indicxnli}
Divyanshu Aggarwal, Vivek Gupta, and Anoop Kunchukuttan. 2022.
\newblock \href {https://doi.org/10.18653/v1/2022.emnlp-main.755} {{I}ndic{XNLI}: Evaluating multilingual inference for {I}ndian languages}.
\newblock In \emph{Proceedings of the 2022 Conference on Empirical Methods in Natural Language Processing}, pages 10994--11006, Abu Dhabi, United Arab Emirates. Association for Computational Linguistics.

\bibitem[{Alves et~al.(2024)Alves, Pombal, Guerreiro, Martins, Alves, Farajian, Peters, Rei, Fernandes, Agrawal, Colombo, de~Souza, and Martins}]{alves2024tower}
Duarte~M. Alves, José Pombal, Nuno~M. Guerreiro, Pedro~H. Martins, João Alves, Amin Farajian, Ben Peters, Ricardo Rei, Patrick Fernandes, Sweta Agrawal, Pierre Colombo, José G.~C. de~Souza, and André F.~T. Martins. 2024.
\newblock \href {https://arxiv.org/abs/2402.17733} {Tower: An open multilingual large language model for translation-related tasks}.
\newblock \emph{Preprint}, arXiv:2402.17733.

\bibitem[{Ansell et~al.(2023)Ansell, Parovi{\'c}, Vuli{\'c}, Korhonen, and Ponti}]{ansell-etal-2023-unifying}
Alan Ansell, Marinela Parovi{\'c}, Ivan Vuli{\'c}, Anna Korhonen, and Edoardo Ponti. 2023.
\newblock \href {https://doi.org/10.18653/v1/2023.emnlp-main.242} {Unifying cross-lingual transfer across scenarios of resource scarcity}.
\newblock In \emph{Proceedings of the 2023 Conference on Empirical Methods in Natural Language Processing}, pages 3980--3995, Singapore. Association for Computational Linguistics.

\bibitem[{Ansell et~al.(2022)Ansell, Ponti, Korhonen, and Vuli{\'c}}]{ansell-etal-2022-composable}
Alan Ansell, Edoardo Ponti, Anna Korhonen, and Ivan Vuli{\'c}. 2022.
\newblock \href {https://doi.org/10.18653/v1/2022.acl-long.125} {Composable sparse fine-tuning for cross-lingual transfer}.
\newblock In \emph{Proceedings of the 60th Annual Meeting of the Association for Computational Linguistics (Volume 1: Long Papers)}, pages 1778--1796, Dublin, Ireland. Association for Computational Linguistics.

\bibitem[{Artetxe et~al.(2020)Artetxe, Ruder, and Yogatama}]{artetxe-etal-2020-cross}
Mikel Artetxe, Sebastian Ruder, and Dani Yogatama. 2020.
\newblock \href {https://doi.org/10.18653/v1/2020.acl-main.421} {On the cross-lingual transferability of monolingual representations}.
\newblock In \emph{Proceedings of the 58th Annual Meeting of the Association for Computational Linguistics}, pages 4623--4637, Online. Association for Computational Linguistics.

\bibitem[{Bapna and Firat(2019)}]{bapna-firat-2019-simple}
Ankur Bapna and Orhan Firat. 2019.
\newblock \href {https://doi.org/10.18653/v1/D19-1165} {Simple, scalable adaptation for neural machine translation}.
\newblock In \emph{Proceedings of the 2019 Conference on Empirical Methods in Natural Language Processing and the 9th International Joint Conference on Natural Language Processing (EMNLP-IJCNLP)}, pages 1538--1548, Hong Kong, China. Association for Computational Linguistics.

\bibitem[{Baziotis et~al.(2022)Baziotis, Artetxe, Cross, and Bhosale}]{baziotis-etal-2022-multilingual}
Christos Baziotis, Mikel Artetxe, James Cross, and Shruti Bhosale. 2022.
\newblock \href {https://doi.org/10.18653/v1/2022.emnlp-main.77} {Multilingual machine translation with hyper-adapters}.
\newblock In \emph{Proceedings of the 2022 Conference on Empirical Methods in Natural Language Processing}, pages 1170--1185, Abu Dhabi, United Arab Emirates. Association for Computational Linguistics.

\bibitem[{Chronopoulou et~al.(2023{\natexlab{a}})Chronopoulou, Peters, Fraser, and Dodge}]{chronopoulou-etal-2023-adaptersoup}
Alexandra Chronopoulou, Matthew Peters, Alexander Fraser, and Jesse Dodge. 2023{\natexlab{a}}.
\newblock \href {https://doi.org/10.18653/v1/2023.findings-eacl.153} {{A}dapter{S}oup: Weight averaging to improve generalization of pretrained language models}.
\newblock In \emph{Findings of the Association for Computational Linguistics: EACL 2023}, pages 2054--2063, Dubrovnik, Croatia. Association for Computational Linguistics.

\bibitem[{Chronopoulou et~al.(2023{\natexlab{b}})Chronopoulou, Stojanovski, and Fraser}]{chronopoulou-etal-2023-language}
Alexandra Chronopoulou, Dario Stojanovski, and Alexander Fraser. 2023{\natexlab{b}}.
\newblock \href {https://doi.org/10.18653/v1/2023.loresmt-1.5} {Language-family adapters for low-resource multilingual neural machine translation}.
\newblock In \emph{Proceedings of the The Sixth Workshop on Technologies for Machine Translation of Low-Resource Languages (LoResMT 2023)}, pages 59--72, Dubrovnik, Croatia. Association for Computational Linguistics.

\bibitem[{Conneau et~al.(2020)Conneau, Khandelwal, Goyal, Chaudhary, Wenzek, Guzm{\'a}n, Grave, Ott, Zettlemoyer, and Stoyanov}]{conneau-etal-2020-unsupervised}
Alexis Conneau, Kartikay Khandelwal, Naman Goyal, Vishrav Chaudhary, Guillaume Wenzek, Francisco Guzm{\'a}n, Edouard Grave, Myle Ott, Luke Zettlemoyer, and Veselin Stoyanov. 2020.
\newblock \href {https://doi.org/10.18653/v1/2020.acl-main.747} {Unsupervised cross-lingual representation learning at scale}.
\newblock In \emph{Proceedings of the 58th Annual Meeting of the Association for Computational Linguistics}, pages 8440--8451, Online. Association for Computational Linguistics.

\bibitem[{Conneau et~al.(2018)Conneau, Rinott, Lample, Williams, Bowman, Schwenk, and Stoyanov}]{conneau2018xnli}
Alexis Conneau, Ruty Rinott, Guillaume Lample, Adina Williams, Samuel Bowman, Holger Schwenk, and Veselin Stoyanov. 2018.
\newblock \href {https://doi.org/10.18653/v1/D18-1269} {{XNLI}: Evaluating cross-lingual sentence representations}.
\newblock In \emph{Proceedings of the 2018 Conference on Empirical Methods in Natural Language Processing}, pages 2475--2485, Brussels, Belgium. Association for Computational Linguistics.

\bibitem[{Devlin et~al.(2019)Devlin, Chang, Lee, and Toutanova}]{devlin-etal-2019-bert}
Jacob Devlin, Ming-Wei Chang, Kenton Lee, and Kristina Toutanova. 2019.
\newblock \href {https://doi.org/10.18653/v1/N19-1423} {{BERT}: Pre-training of deep bidirectional transformers for language understanding}.
\newblock In \emph{Proceedings of the 2019 Conference of the North {A}merican Chapter of the Association for Computational Linguistics: Human Language Technologies, Volume 1 (Long and Short Papers)}, pages 4171--4186, Minneapolis, Minnesota. Association for Computational Linguistics.

\bibitem[{Gururangan et~al.(2020)Gururangan, Marasovi{\'c}, Swayamdipta, Lo, Beltagy, Downey, and Smith}]{gururangan-etal-2020-dont}
Suchin Gururangan, Ana Marasovi{\'c}, Swabha Swayamdipta, Kyle Lo, Iz~Beltagy, Doug Downey, and Noah~A. Smith. 2020.
\newblock \href {https://doi.org/10.18653/v1/2020.acl-main.740} {Don{'}t stop pretraining: Adapt language models to domains and tasks}.
\newblock In \emph{Proceedings of the 58th Annual Meeting of the Association for Computational Linguistics}, pages 8342--8360, Online. Association for Computational Linguistics.

\bibitem[{Ha et~al.(2017)Ha, Dai, and Le}]{ha2017hypernetworks}
David Ha, Andrew~M. Dai, and Quoc~V. Le. 2017.
\newblock Hypernetworks.
\newblock In \emph{International Conference on Learning Representations}.

\bibitem[{Haddow et~al.(2022)Haddow, Bawden, Miceli~Barone, Helcl, and Birch}]{haddow-etal-2022-survey}
Barry Haddow, Rachel Bawden, Antonio~Valerio Miceli~Barone, Jind{\v{r}}ich Helcl, and Alexandra Birch. 2022.
\newblock \href {https://doi.org/10.1162/coli_a_00446} {Survey of low-resource machine translation}.
\newblock \emph{Computational Linguistics}, 48(3):673--732.

\bibitem[{Houlsby et~al.(2019)Houlsby, Giurgiu, Jastrzebski, Morrone, De~Laroussilhe, Gesmundo, Attariyan, and Gelly}]{pmlr-v97-houlsby19a}
Neil Houlsby, Andrei Giurgiu, Stanislaw Jastrzebski, Bruna Morrone, Quentin De~Laroussilhe, Andrea Gesmundo, Mona Attariyan, and Sylvain Gelly. 2019.
\newblock Parameter-efficient transfer learning for {NLP}.
\newblock In \emph{Proceedings of the 36th International Conference on Machine Learning}, volume~97 of \emph{Proceedings of Machine Learning Research}, pages 2790--2799. PMLR.

\bibitem[{Huang et~al.(2024)Huang, Liu, Lin, Pang, Du, and Lin}]{huang2024lorahub}
Chengsong Huang, Qian Liu, Bill~Yuchen Lin, Tianyu Pang, Chao Du, and Min Lin. 2024.
\newblock \href {https://arxiv.org/abs/2307.13269} {Lorahub: Efficient cross-task generalization via dynamic lora composition}.
\newblock \emph{Preprint}, arXiv:2307.13269.

\bibitem[{Ilharco et~al.(2023)Ilharco, Ribeiro, Wortsman, Schmidt, Hajishirzi, and Farhadi}]{DBLP:conf/iclr/IlharcoRWSHF23}
Gabriel Ilharco, Marco~T{\'{u}}lio Ribeiro, Mitchell Wortsman, Ludwig Schmidt, Hannaneh Hajishirzi, and Ali Farhadi. 2023.
\newblock Editing models with task arithmetic.
\newblock In \emph{The Eleventh International Conference on Learning Representations, {ICLR} 2023, Kigali, Rwanda, May 1-5, 2023}. OpenReview.net.

\bibitem[{Klimaszewski et~al.(2023)Klimaszewski, Belligoli, Kumar, and Stergiadis}]{klimaszewski23}
Mateusz Klimaszewski, Zeno Belligoli, Satendra Kumar, and Emmanouil Stergiadis. 2023.
\newblock \href {https://doi.org/10.3233/FAIA230404} {Gated adapters for multi-domain neural machine translation}.
\newblock In \emph{{ECAI} 2023 - 26th European Conference on Artificial Intelligence}, volume 372 of \emph{Frontiers in Artificial Intelligence and Applications}, pages 1264--1271. {IOS} Press.

\bibitem[{Kocmi et~al.(2021)Kocmi, Mach{\'{a}}{\v{c}}ek, and Bojar}]{kocmi2022reality}
Tom Kocmi, Dominik Mach{\'{a}}{\v{c}}ek, and Ond{\v{r}}ej Bojar. 2021.
\newblock \emph{The Reality of Multi-Lingual Machine Translation}, volume~21 of \emph{Studies in Computational and Theoretical Linguistics}.
\newblock Institute of Formal and Applied Linguistics, Prague, Czechia.

\bibitem[{Lee et~al.(2022)Lee, Hwang, and Kim}]{lee-etal-2022-fad}
Jaeseong Lee, Seung-won Hwang, and Taesup Kim. 2022.
\newblock {FAD}-{X}: Fusing adapters for cross-lingual transfer to low-resource languages.
\newblock In \emph{Proceedings of the 2nd Conference of the Asia-Pacific Chapter of the Association for Computational Linguistics and the 12th International Joint Conference on Natural Language Processing (Volume 2: Short Papers)}, pages 57--64, Online only. Association for Computational Linguistics.

\bibitem[{Littell et~al.(2017)Littell, Mortensen, Lin, Kairis, Turner, and Levin}]{littell-etal-2017-uriel}
Patrick Littell, David~R. Mortensen, Ke~Lin, Katherine Kairis, Carlisle Turner, and Lori Levin. 2017.
\newblock \href {https://aclanthology.org/E17-2002} {{URIEL} and lang2vec: Representing languages as typological, geographical, and phylogenetic vectors}.
\newblock In \emph{Proceedings of the 15th Conference of the {E}uropean Chapter of the Association for Computational Linguistics: Volume 2, Short Papers}, pages 8--14, Valencia, Spain. Association for Computational Linguistics.

\bibitem[{Malaviya et~al.(2017)Malaviya, Neubig, and Littell}]{malaviya-etal-2017-learning}
Chaitanya Malaviya, Graham Neubig, and Patrick Littell. 2017.
\newblock \href {https://doi.org/10.18653/v1/D17-1268} {Learning language representations for typology prediction}.
\newblock In \emph{Proceedings of the 2017 Conference on Empirical Methods in Natural Language Processing}, pages 2529--2535, Copenhagen, Denmark. Association for Computational Linguistics.

\bibitem[{Martin et~al.(2020)Martin, Muller, Ortiz~Su{\'a}rez, Dupont, Romary, de~la Clergerie, Seddah, and Sagot}]{martin-etal-2020-camembert}
Louis Martin, Benjamin Muller, Pedro~Javier Ortiz~Su{\'a}rez, Yoann Dupont, Laurent Romary, {\'E}ric de~la Clergerie, Djam{\'e} Seddah, and Beno{\^\i}t Sagot. 2020.
\newblock \href {https://doi.org/10.18653/v1/2020.acl-main.645} {{C}amem{BERT}: a tasty {F}rench language model}.
\newblock In \emph{Proceedings of the 58th Annual Meeting of the Association for Computational Linguistics}, pages 7203--7219, Online. Association for Computational Linguistics.

\bibitem[{Muennighoff et~al.(2023)Muennighoff, Wang, Sutawika, Roberts, Biderman, Le~Scao, Bari, Shen, Yong, Schoelkopf, Tang, Radev, Aji, Almubarak, Albanie, Alyafeai, Webson, Raff, and Raffel}]{muennighoff-etal-2023-crosslingual}
Niklas Muennighoff, Thomas Wang, Lintang Sutawika, Adam Roberts, Stella Biderman, Teven Le~Scao, M~Saiful Bari, Sheng Shen, Zheng~Xin Yong, Hailey Schoelkopf, Xiangru Tang, Dragomir Radev, Alham~Fikri Aji, Khalid Almubarak, Samuel Albanie, Zaid Alyafeai, Albert Webson, Edward Raff, and Colin Raffel. 2023.
\newblock \href {https://doi.org/10.18653/v1/2023.acl-long.891} {Crosslingual generalization through multitask finetuning}.
\newblock In \emph{Proceedings of the 61st Annual Meeting of the Association for Computational Linguistics (Volume 1: Long Papers)}, pages 15991--16111, Toronto, Canada. Association for Computational Linguistics.

\bibitem[{Nozza et~al.(2020)Nozza, Bianchi, and Hovy}]{Nozza2020WhatT}
Debora Nozza, Federico Bianchi, and Dirk Hovy. 2020.
\newblock What the [mask]? making sense of language-specific bert models.
\newblock \emph{ArXiv}, abs/2003.02912.

\bibitem[{Parovi{\'c} et~al.(2022)Parovi{\'c}, Glava{\v{s}}, Vuli{\'c}, and Korhonen}]{parovic-etal-2022-bad}
Marinela Parovi{\'c}, Goran Glava{\v{s}}, Ivan Vuli{\'c}, and Anna Korhonen. 2022.
\newblock \href {https://doi.org/10.18653/v1/2022.naacl-main.130} {{BAD}-{X}: Bilingual adapters improve zero-shot cross-lingual transfer}.
\newblock In \emph{Proceedings of the 2022 Conference of the North American Chapter of the Association for Computational Linguistics: Human Language Technologies}, pages 1791--1799, Seattle, United States. Association for Computational Linguistics.

\bibitem[{Parovi{\'c} et~al.(2024)Parovi{\'c}, Vuli{\'c}, and Korhonen}]{parovic-etal-2024-investigating}
Marinela Parovi{\'c}, Ivan Vuli{\'c}, and Anna Korhonen. 2024.
\newblock Investigating the potential of task arithmetic for cross-lingual transfer.
\newblock In \emph{Proceedings of the 18th Conference of the European Chapter of the Association for Computational Linguistics (Volume 2: Short Papers)}, pages 124--137, St. Julian{'}s, Malta. Association for Computational Linguistics.

\bibitem[{Pfeiffer et~al.(2022)Pfeiffer, Goyal, Lin, Li, Cross, Riedel, and Artetxe}]{pfeiffer-etal-2022-lifting}
Jonas Pfeiffer, Naman Goyal, Xi~Lin, Xian Li, James Cross, Sebastian Riedel, and Mikel Artetxe. 2022.
\newblock \href {https://doi.org/10.18653/v1/2022.naacl-main.255} {Lifting the curse of multilinguality by pre-training modular transformers}.
\newblock In \emph{Proceedings of the 2022 Conference of the North American Chapter of the Association for Computational Linguistics: Human Language Technologies}, pages 3479--3495, Seattle, United States. Association for Computational Linguistics.

\bibitem[{Pfeiffer et~al.(2021)Pfeiffer, Kamath, R{\"u}ckl{\'e}, Cho, and Gurevych}]{pfeiffer-etal-2021-adapterfusion}
Jonas Pfeiffer, Aishwarya Kamath, Andreas R{\"u}ckl{\'e}, Kyunghyun Cho, and Iryna Gurevych. 2021.
\newblock \href {https://doi.org/10.18653/v1/2021.eacl-main.39} {{A}dapter{F}usion: Non-destructive task composition for transfer learning}.
\newblock In \emph{Proceedings of the 16th Conference of the European Chapter of the Association for Computational Linguistics: Main Volume}, pages 487--503, Online. Association for Computational Linguistics.

\bibitem[{Pfeiffer et~al.(2020{\natexlab{a}})Pfeiffer, R{\"u}ckl{\'e}, Poth, Kamath, Vuli{\'c}, Ruder, Cho, and Gurevych}]{pfeiffer-etal-2020-adapterhub}
Jonas Pfeiffer, Andreas R{\"u}ckl{\'e}, Clifton Poth, Aishwarya Kamath, Ivan Vuli{\'c}, Sebastian Ruder, Kyunghyun Cho, and Iryna Gurevych. 2020{\natexlab{a}}.
\newblock \href {https://doi.org/10.18653/v1/2020.emnlp-demos.7} {{A}dapter{H}ub: A framework for adapting transformers}.
\newblock In \emph{Proceedings of the 2020 Conference on Empirical Methods in Natural Language Processing: System Demonstrations}, pages 46--54, Online. Association for Computational Linguistics.

\bibitem[{Pfeiffer et~al.(2023)Pfeiffer, Ruder, Vuli{\'c}, and Ponti}]{pfeiffer2023modular}
Jonas Pfeiffer, Sebastian Ruder, Ivan Vuli{\'c}, and Edoardo Ponti. 2023.
\newblock Modular deep learning.
\newblock \emph{Transactions on Machine Learning Research}.
\newblock Survey Certification.

\bibitem[{Pfeiffer et~al.(2020{\natexlab{b}})Pfeiffer, Vuli{\'c}, Gurevych, and Ruder}]{pfeiffer-etal-2020-mad}
Jonas Pfeiffer, Ivan Vuli{\'c}, Iryna Gurevych, and Sebastian Ruder. 2020{\natexlab{b}}.
\newblock \href {https://doi.org/10.18653/v1/2020.emnlp-main.617} {{MAD-X}: {A}n {A}dapter-{B}ased {F}ramework for {M}ulti-{T}ask {C}ross-{L}ingual {T}ransfer}.
\newblock In \emph{Proceedings of the 2020 Conference on Empirical Methods in Natural Language Processing (EMNLP)}, pages 7654--7673, Online. Association for Computational Linguistics.

\bibitem[{Pham et~al.(2020)Pham, Crego, Yvon, and Senellart}]{pham-etal-2020-study}
Minh~Quang Pham, Josep~Maria Crego, Fran{\c{c}}ois Yvon, and Jean Senellart. 2020.
\newblock A study of residual adapters for multi-domain neural machine translation.
\newblock In \emph{Proceedings of the Fifth Conference on Machine Translation}, pages 617--628, Online. Association for Computational Linguistics.

\bibitem[{Philip et~al.(2020)Philip, Berard, Gall{\'e}, and Besacier}]{philip-etal-2020-monolingual}
Jerin Philip, Alexandre Berard, Matthias Gall{\'e}, and Laurent Besacier. 2020.
\newblock \href {https://doi.org/10.18653/v1/2020.emnlp-main.361} {Monolingual adapters for zero-shot neural machine translation}.
\newblock In \emph{Proceedings of the 2020 Conference on Empirical Methods in Natural Language Processing (EMNLP)}, pages 4465--4470, Online. Association for Computational Linguistics.

\bibitem[{Rahimi et~al.(2019)Rahimi, Li, and Cohn}]{rahimi-etal-2019-massively}
Afshin Rahimi, Yuan Li, and Trevor Cohn. 2019.
\newblock Massively multilingual transfer for {NER}.
\newblock In \emph{Proceedings of the 57th Annual Meeting of the Association for Computational Linguistics}, pages 151--164, Florence, Italy. Association for Computational Linguistics.

\bibitem[{Rajpurkar et~al.(2016)Rajpurkar, Zhang, Lopyrev, and Liang}]{rajpurkar-etal-2016-squad}
Pranav Rajpurkar, Jian Zhang, Konstantin Lopyrev, and Percy Liang. 2016.
\newblock \href {https://doi.org/10.18653/v1/D16-1264} {{SQ}u{AD}: 100,000+ questions for machine comprehension of text}.
\newblock In \emph{Proceedings of the 2016 Conference on Empirical Methods in Natural Language Processing}, pages 2383--2392, Austin, Texas. Association for Computational Linguistics.

\bibitem[{Rebuffi et~al.(2017)Rebuffi, Bilen, and Vedaldi}]{NIPS2017_e7b24b11}
Sylvestre-Alvise Rebuffi, Hakan Bilen, and Andrea Vedaldi. 2017.
\newblock Learning multiple visual domains with residual adapters.
\newblock In \emph{Advances in Neural Information Processing Systems}, volume~30. Curran Associates, Inc.

\bibitem[{Rei et~al.(2020)Rei, Stewart, Farinha, and Lavie}]{rei-etal-2020-comet}
Ricardo Rei, Craig Stewart, Ana~C Farinha, and Alon Lavie. 2020.
\newblock \href {https://doi.org/10.18653/v1/2020.emnlp-main.213} {{COMET}: A neural framework for {MT} evaluation}.
\newblock In \emph{Proceedings of the 2020 Conference on Empirical Methods in Natural Language Processing (EMNLP)}, pages 2685--2702, Online. Association for Computational Linguistics.

\bibitem[{Stoica et~al.(2023)Stoica, Bolya, Bjorner, Hearn, and Hoffman}]{stoica2023zipit}
George Stoica, Daniel Bolya, Jakob Bjorner, Taylor Hearn, and Judy Hoffman. 2023.
\newblock \href {https://arxiv.org/abs/2305.03053} {Zipit! merging models from different tasks without training}.
\newblock \emph{Preprint}, arXiv:2305.03053.

\bibitem[{{\"U}st{\"u}n et~al.(2020){\"U}st{\"u}n, Bisazza, Bouma, and van Noord}]{ustun-etal-2020-udapter}
Ahmet {\"U}st{\"u}n, Arianna Bisazza, Gosse Bouma, and Gertjan van Noord. 2020.
\newblock \href {https://doi.org/10.18653/v1/2020.emnlp-main.180} {{UD}apter: Language adaptation for truly {U}niversal {D}ependency parsing}.
\newblock In \emph{Proceedings of the 2020 Conference on Empirical Methods in Natural Language Processing (EMNLP)}, pages 2302--2315, Online. Association for Computational Linguistics.

\bibitem[{Wang et~al.(2023)Wang, Chen, Zhang, Hu, Xu, and Zheng}]{wang-etal-2023-adapterdistillation}
Junjie Wang, Yicheng Chen, Wangshu Zhang, Sen Hu, Teng Xu, and Jing Zheng. 2023.
\newblock \href {https://doi.org/10.18653/v1/2023.emnlp-industry.20} {{A}dapter{D}istillation: Non-destructive task composition with knowledge distillation}.
\newblock In \emph{Proceedings of the 2023 Conference on Empirical Methods in Natural Language Processing: Industry Track}, pages 194--201, Singapore. Association for Computational Linguistics.

\bibitem[{Wang et~al.(2020)Wang, Lipton, and Tsvetkov}]{wang-etal-2020-negative}
Zirui Wang, Zachary~C. Lipton, and Yulia Tsvetkov. 2020.
\newblock \href {https://doi.org/10.18653/v1/2020.emnlp-main.359} {On negative interference in multilingual models: Findings and a meta-learning treatment}.
\newblock In \emph{Proceedings of the 2020 Conference on Empirical Methods in Natural Language Processing (EMNLP)}, pages 4438--4450, Online. Association for Computational Linguistics.

\bibitem[{Workshop(2023)}]{workshop2023bloom}
BigScience Workshop. 2023.
\newblock \href {https://arxiv.org/abs/2211.05100} {Bloom: A 176b-parameter open-access multilingual language model}.
\newblock \emph{Preprint}, arXiv:2211.05100.

\bibitem[{Yadav et~al.(2023)Yadav, Tam, Choshen, Raffel, and Bansal}]{yadav2023tiesmerging}
Prateek Yadav, Derek Tam, Leshem Choshen, Colin Raffel, and Mohit Bansal. 2023.
\newblock {TIES}-merging: Resolving interference when merging models.
\newblock In \emph{Thirty-seventh Conference on Neural Information Processing Systems}.

\bibitem[{Zhang et~al.(2023)Zhang, Chen, Liu, and He}]{zhang2023composing}
Jinghan Zhang, Shiqi Chen, Junteng Liu, and Junxian He. 2023.
\newblock Composing parameter-efficient modules with arithmetic operation.
\newblock In \emph{Thirty-seventh Conference on Neural Information Processing Systems}.

\end{thebibliography}

\appendix

\section{Related languages}
\label{sec:appendix_related}
We present the list of related languages used in our experiments in Table \ref{tab:related} (details in Section \ref{para:related}).

\begin{table}[b]
    \centering
    \begin{tabular}{lccccccc}
         \toprule
         Lang. & ar & bg & de & el & es & fr & hi \\
         Related & sw & ru & fr & es & fr & es & ur \\
         \hdashline
         Lang. & ru & sw & tr & ur & vi & zh & \\
         Related & bg & ar & bg & hi & ru & ar & \\
         \bottomrule
    \end{tabular}
    \caption{\label{tab:related}Languages used in the experiments with corresponding related languages. Details are provided in Section \ref{para:related}.}
\end{table}

\section{Zero-shot evaluation}

\begin{figure}[t]
    \centering
    \includesvg[width=0.88\linewidth]{figures/NER_mBERT_zeroshot.svg}

    \vspace{-0.25cm}

    \includesvg[width=0.88\linewidth]{figures/NLI_mBERT_zeroshot.svg}

    \vspace{-0.25cm}

    \includesvg[width=0.88\linewidth]{figures/QA_mBERT_zeroshot.svg}
    
    \caption{Zero-shot mBERT language arithmetic evaluation, where one side of the arithmetic is an English adapter, and the other is related to the target language adapter (e.g. French for Spanish - $LA_{es}(en, fr)$). The values above bars present a relative difference to a better proxy.}
    \label{fig:mBERT_zeroshot}
\end{figure}

Figure \ref{fig:mBERT_zeroshot} presents the results of the experiments described in Section \ref{sec:zero_shot} for the mBERT model.

\section{Improving existing language adapters}
Figure \ref{fig:mbert_improving} presents the results of the experiments described in Section \ref{sec:improving} for the mBERT model.

\begin{figure}[t]
    \centering
    \includesvg[width=0.88\linewidth]{figures/NER_mBERT.svg}

    \vspace{-0.25cm}

    \includesvg[width=0.88\linewidth]
    {figures/NLI_mBERT.svg}

    \vspace{-0.25cm}

    \includesvg[width=0.88\linewidth]{figures/QA_mBERT.svg}
    
    \caption{Variants of language arithmetic compared to the MAD-X method in the use-case to improve an existing target language adapter. The values above bars present a difference between a better LA setup and the MAD-X framework for the mBERT model.}
    \label{fig:mbert_improving}
\end{figure}

\section{Lambda impact - NLI and QA}
\label{sec:appendix_lambda_nli_qa}

Figure \ref{fig:zeroshot_interpolation_nli_qa} presents the analysis of lambda impact for NLI and QA tasks. For details, refer to Section \ref{sec:analysis_lambda}.

\begin{figure}[t]
  \centering
  \includesvg[width=0.9\linewidth]{figures/zeroshot_nli_xlmr}
  \centering
  \includesvg[width=0.9\linewidth]{figures/zeroshot_qa_xlmr}
\caption{Interpolation of $\lambda$ values for the zero-shot NLI and QA XLM-R scenario on the validation dataset. The horizontal dashed lines represent the baseline scores for both languages used in language arithmetic.}
\label{fig:zeroshot_interpolation_nli_qa}
\end{figure}

\section{Language vs task vectors}
\label{sec:lang_vs_task}

Task vectors exhibit high sparsity and orthogonality, as \citet{DBLP:conf/iclr/IlharcoRWSHF23} observed. While the former characteristic can be denoted in language vectors (Figure \ref{fig:sparsity}), the latter displays different properties, in contrast to task vectors. In Figure~\ref{fig:cosine_sim}, we visualise the cosine similarity between evaluated language vectors of language adapters. Notably, the minimal cosine similarity ($0.19$) surpasses the maximum ($0.18$) reported by previous research in the task space \cite{DBLP:conf/iclr/IlharcoRWSHF23}. Interestingly, most pairs in the task space oscillate within the range of $0.01$ to $0.03$. At the same time, language vectors surpass $0.2$ in almost each case, indicating that the orthogonality aspect is an inherent property of task adapters.

\begin{figure}
    \centering
    \includesvg[width=0.6\linewidth]{figures/sparsity.svg}
    \caption{
    Language vectors, similar to task vectors, are extremely sparse. The kernel density estimate plot presents the weights of a Spanish mBERT adapter. The behaviour is consistent across sampled layers and languages.
    }
    \label{fig:sparsity}
\end{figure}

Based on the cosine similarity observation, we investigated one of the recent task arithmetic extensions, Ties-Merging \cite{yadav2023tiesmerging}. This work introduces a three-step algorithm that prevents different parameter interferences, improving upon task arithmetic. The algorithm \textit{decreases the cosine similarity} via a pruning step and alignment of parameter signs to perform arithmetic only on relevant parameters to the merged tasks. On the experimental details note, as Ties-Merging operates on averaging, not addition, we utilise a different lambda range during validation (as suggested by \citet{yadav2023tiesmerging},  $\lambda \in [0.8, 1.8]$), and we set Top-K\% to the default value of $20$ and additionally to $80$.

\begin{figure}[t]
    \centering
    \includesvg[width=\linewidth]{figures/cosine_sim_matrix}
    \caption{Cosine similarity between language vectors of language adapters.}
    \label{fig:cosine_sim}
\end{figure}

We report the comparison in the zero-shot setting on the NER task (XLM-R version) in Table \ref{tab:ties}. The Ties-Merging decreases the results significantly compared to the default language arithmetic. Moreover, we note that the pruning operation has the reverse effect; higher pruning (i.e. keeping Top-K\% lower)  decreases the performance  (in contrast to task vectors) by making language vectors more sparse and, hence, closer to orthogonal.

\begin{table}
    \centering
    \begin{tabular}{lc}
        \toprule
        Method & AVG F1 score \\
        \midrule
         LA & 60.54 \\
         Ties-Merging (Top-K\% 20) & 52.94 \\
         Ties-Merging (Top-K\% 80) & 57.57 \\
         \bottomrule
    \end{tabular}
    \caption{Ties-Merging evaluation in the zero-shot setup on the NER task (XLM-R version, averaged over three runs and all evaluated languages). In the case of language arithmetic, where the language vectors have a higher overlap (i.e. higher cosine similarity), removing parameter interference decreases the overall performance.}
    \label{tab:ties}
\end{table}

One interpretation of the phenomena can be the different goals of the arithmetic: in the multi-task setup, we try to include multiple, often disconnected, tasks into a single task vector. In contrast, the language vectors' goal is to include the knowledge of the closely related language rather than remove the harmful artefacts. Our findings indicate that language arithmetic has different characteristics than task arithmetic, and the follow-up works that improve upon task arithmetic might not be suited for the multilingual context.

\end{document}